\documentclass[Conference]{IEEEtran}
\usepackage[utf8]{inputenc}
\usepackage[english]{babel}
\usepackage{amsmath}
\usepackage{amsfonts}
\usepackage{amssymb}
\usepackage{graphicx}
 
\usepackage{amsthm}
\usepackage{algorithm}
\usepackage{algpseudocode, setspace}
\usepackage{float}
\usepackage{xcolor}
\usepackage{multirow}
\usepackage{lipsum}
\usepackage{subcaption}

\author{Rugved katole$^{1}$, Arpita Sinha$^{1}$% <-this % stops a space
\thanks{$^{1}$The authors belong to the Systems and Control Engineering department, Indian Institute Of Technology, Bombay
        }%
}

\title{A low-cost Framework for Decentralized Autonomous Intersection Management}
\begin{document}
\maketitle
\begin{abstract}
This paper addresses the traffic management problem for autonomous vehicles at intersections without traffic signals. In the current system, a road junction has no traffic signals when the traffic volume is low to medium. Installing infrastructure at each unsignalled crossing to coordinate autonomous cars can be formidable. We propose a novel low-cost solution strategy where the vehicles use a harmony matrix to find the best possible combination of the cars to cross the intersection without any crashes. The harmony matrix defines the connection between different vehicle maneuvers and is queried online for intersection management.
We maximize the throughput of the intersection by solving a maximal clique problem formulated based on the vehicles present at the intersection. The proposed algorithm relies on the intent perceived by the autonomous vehicles. We compare our work with a communication-based strategy that uses V2I communication protocols, and through extensive simulation, we showed that our algorithm is comparable when the traffic volume is less than 500 PCUs/hr/lane.
\end{abstract}

\section{Introduction}
Efficient management of unsignalized intersections is crucial for smooth, undisrupted traffic flow. An inadequately managed unsignalized intersection can affect the network's signalized intersections as well as intelligent transportation systems (ITS) and may eventually lead to congestion and road accidents. In accordance with the data provided by the Ministry of Road Transport and Highways, India (MORTH), in 2021, a staggering 98,571 accidents were related to intersections, with a vast majority of 74.21\% (i.e., 73,155) occurring at uncontrolled intersections \cite{TransportResearchWing2021RoadIndia}. Similarly, in the EU, 20\% of fatalities are intersection-related, and in the USA, 21.5\% of intersection crashes lead to fatalities out of a total of 40\% \cite{Chen2016CooperativeSurvey}. The data suggests intersections account for the majority of road accidents, highlighting the urgent need for innovative intersection management solutions like Autonomous Intersection Management (AIM) \cite{2010CrashPerspective}.

In many towns and cities, uncontrolled intersections lacking proper infrastructure are common, particularly in areas with low traffic density. The absence of infrastructure can cause traffic flow disruptions, leading to safety issues. To mitigate these problems, the installation of the required infrastructure is imperative. As autonomous vehicles become more accessible, it is possible that intersections will be managed autonomously through a command center in the near future. Nevertheless, even with technological advancements, some infrastructure, such as a traditional traffic signal or a command center, will still be necessary for uncontrolled intersections. However, from an economic standpoint, building such infrastructure may come with exorbitant costs that may not justify its utility. 

Given the potential risks and costs associated with uncontrolled intersections, it is crucial to find effective ways of managing them. This question of intersection management becomes even more pressing with the increasing prevalence of autonomous vehicles. However, it remains unclear how such vehicles can navigate uncontrolled intersections without the aid of infrastructure or communication protocols like V2X/V2I/V2V (vehicle-to-everything/vehicle-to-infrastructure/vehicle-to-vehicle).

Traditionally, traffic lights have been a commonly used tool for regulating traffic flow at intersections for many years. In most cases, the signal phase timings are determined using various parameters such as traffic density (PCUs/hr) and queue length. For instance, the Traffic Signal Timing manual by NCHRP provides guidelines for optimizing signal timings based on these factors \cite{Urbanik2015SignalEdition}. However, despite their widespread use, traffic lights have been shown to be inefficient in minimizing waiting times at intersections. Prolonged waiting times lead to increased fuel consumption, resulting in higher levels of harmful emissions. On the contrary, Adaptive traffic lights utilize data collection technologies (viz. connected autonomous vehicles (CAVs), mobile sensing, etc.) to gather real-time traffic data and optimize traffic phase timings based on parameters such as flow volume, travel time, queue length, and shockwave boundary \cite{Guo2019UrbanSurvey}. However, it costs around 65,000 USD to install an adaptive traffic control system, which makes them a less cost-effective solution in low-traffic areas \cite{Stevanovic2010AdaptivePractice}.

Intelligent traffic control systems with communication infrastructure emerge as a cost-effective solution relative to adaptive traffic lights (approx 7000-8000 USD). A roadside infrastructure maintains the coordination between autonomous vehicles through space-time reservation based on First-Come-First-Serve (FCFS) policy \cite{Dresner2008AManagement}, maximum arrival time \cite{Li2020IntersectionCommunication} or by optimizing trajectory and departure sequence \cite{Yang2016IsolatedVehicles}.
However, a failure in roadside infrastructure would lead to systems breakdown and safety hazards. Connected Autonomous Vehicles (CAVs) using vehicle-to-vehicle communication technology are robust against a single point of failure. Through V2V communication, various parameters like speed profile, desired lane, and position are exchanged for coordination \cite{Li2006CooperativeCommunication}, a non-linear model predictive control is used to obtain trajectory from shared trajectories. However, they rely on limited computational resources at the vehicle end and require high communication and computational bandwidth to reach a consensus. These approaches are similar to ours, where vehicles individually perform the calculations, except with our approach, we eliminate the need for broadcasting and heavy computation.

Other intersection management techniques rely only on onboard perception for their decision-making. To make up for uncertainty in perceived sensor data, scholars have used the concept of pseudo-vehicle to solve a POMDP formulation \cite{Zhang2021ImprovedDriving} or using a belief updater along with POMDPs for trajectory planning \cite{Xia2022InteractiveSigns}.
\cite{Nan2022IntentionIntersection} uses a combination of the Hidden Markov Model, Gaussian Mixture Model, and Support Vector Machines for intent prediction. However, POMDPs and game-theoretic methods suffer through the curse of dimensionality, and their computational complexity hinders real-time implementation. \cite{Zyner2020NaturalisticNetworks} employ recurrent neural networks and a mixture density network output layer, along with a clustering algorithm, to predict multi-modal driver trajectories.
The major focus of these algorithms is to safely navigate the intersection without the infrastructure. However, we believe that less attention is given to maximizing the throughput of the intersection in such a scenario.

In this paper, we propose a novel low-cost framework specifically tailored for intersections where installing traffic signals or any other infrastructure is redundant due to low traffic density ($\leq 500 PCUs/hr/lane$ refer section \ref{sec:simultaion}). The algorithm is designed to enable fully autonomous vehicles to navigate through intersections in a fully autonomous environment without relying on any infrastructure. This is achieved by building a graph using a harmony matrix, and the maximal clique of the graph provides the best combination of vehicles to enter the intersection. 

\textbf{Contributions: }
\begin{itemize}
    \item The proposed algorithm uses a cost-effective method for decision-making at unsignaled intersections.
    \item The algorithm unanimously provides the best possible combination of vehicles to maximize the throughput of the intersection.
\end{itemize}

\section{Problem Formulation}
\label{Problem_form}
In settings characterized by low traffic density, such as intersections within remote villages or campus environments, the deployment of sophisticated traffic solutions reliant on infrastructure may not be economically viable. Consequently, there arises a need for cost-effective and straightforward intersection management solutions tailored to such contexts.
Consider an unsignaled intersection with medium to low traffic density. The objective is to autonomously navigate the intersection without relying on sophisticated infrastructure such that maximum number of vehicles can cross simultaneously. The intersection can have multiple connected roads, but to formulate the problem, we focus on a 4-way intersection as depicted in Figure \ref{fig: solution}A; the algorithm's effectiveness for 3-way and 5-way intersections is shown in the simulations. Since the intersection has low traffic density, it is reasonable to assume that each road at the intersection consists of one incoming lane and one outgoing lane. The vehicles approaching the intersection indicate their intentions using indicator lights (left, right, or none for going straight) such that onboard perception can detect the intent. There is ample literature on identifying the intent without communication. For instance, intent can be inferred through the use of POMDPs \cite{Zhang2021ImprovedDriving} or Neural Networks \cite{Zyner2020NaturalisticNetworks, Agrawal2023TrafficCNN}. Hence, we focus on maximizing the throughput of intersections for autonomous vehicles. The algorithm is modular such that any of the intent prediction algorithms can be integrated to maximize the throughput of the intersection.

%%%%%%%%%%%%%%%%%%%%%%%%%%%%%%%%%%%%%%%%%%%%%%%%%

In our study, we operate under the assumption that all vehicles within the system are autonomous. Upon approaching an intersection, each vehicle engages in the observation of the intentions of vehicles occupying other lanes. Subsequently, utilizing the gathered information, the vehicle autonomously determines whether to proceed through the intersection or await clearance. Our proposed algorithm embodies a decentralized approach, wherein each vehicle possesses the capability to independently establish the sequence for navigating the intersection. The specifics of this decision-making strategy are expounded upon in the subsequent section.

In this work, we have assumed that all vehicles are autonomous vehicles. As a car reaches the intersection, it observes the intention of the vehicles in the other lanes if present. Based on the observation, the car makes a decision independently to wait or cross the lane. The proposed algorithm is a decentralized strategy that allows each vehicle to decide the sequence in which they cross the intersection. We detail the decision-making strategy in the next section.

\section{Decentralized Intersection Management}

Vehicles arriving at the intersection at various speeds need to make a unanimous decision to traverse the intersection safely. 
Prior to elucidating the algorithm, it is imperative to outline the lane marking adjustments and modifications to the existing road infrastructure, which facilitate strategic and unanimous decision-making processes.
Three distinct zones—red, yellow, and green—are to be physically implemented on roads through markings to ensure vehicles approach the intersection at a safe speed, thereby mitigating unnecessary ambiguities in decision-making. 
The red zone is designated to prompt vehicles to decelerate to a safer speed, thus reducing the likelihood of fatalities in the event of an accident. Given previous research findings by the Institute for Road Safety Research \cite{SWOV2011TheCrashes}, indicating that 90\% of pedestrians survive collisions at speeds of 20 km/h, we consider this to be a safe speed threshold. Assuming an average speed limit of 40 km/h on single-lane roads and an average deceleration rate of 2 $m/s^2$ for passenger cars, vehicles can come to a complete stop within a distance of 23.14 meters. Incorporating a buffer distance, we propose the length of the red zone to be 30 meters.

The yellow zone is structured to accommodate a single vehicle, serving as a space for observation and participation in decision-making for other vehicles. The ego vehicle remains within the yellow zone until granted the right of way. During this period, it monitors vehicles in the yellow zones of adjacent lanes and continuously evaluates right-of-way conditions using the prescribed algorithm. It is stipulated that the length of the yellow zone must be at least equal to the dimensions of a typical car. Assuming an average passenger car length of 4 meters and a headway distance of 2 meters, we propose a yellow zone length of 6 meters, ensuring that only one vehicle occupies the yellow zone at any given time.

The green zone serves as a safety buffer during intersection crossings to address potential discrepancies in right-of-way decisions. In instances where a new vehicle enters the yellow zone after a right-of-way decision has been made but before preceding vehicles can clear the intersection, the subsequent decision might conflict with prior permissions granted. To mitigate such occurrences, the green zone acts as a buffer, preventing vehicles from entering the intersection under undesired circumstances. Once a vehicle has traversed over half of the green zone, it is granted the right of way by others. The length of the green zone is determined based on the frequency of such interventions. In low-traffic areas characterized by infrequent interventions, the green zone is set to half the length of a passenger car, i.e., $4 \times 0.5 = 2 $m.

\begin{figure}
    \centering
    \includegraphics[width = 3 in]{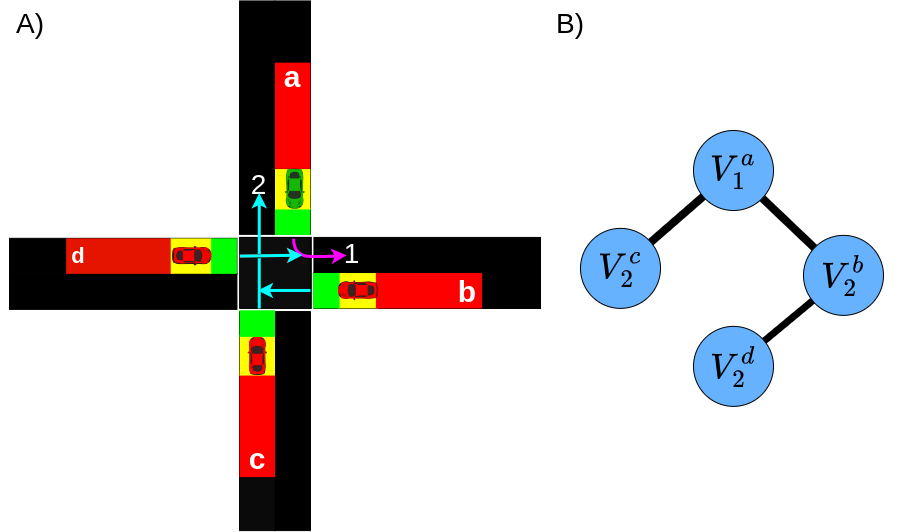}
    \caption{A) A scenario of 4 crossing vehicles with conflicting movements B) A graph generated using a harmony matrix based on vehicle movements}
    \label{fig: solution}
\end{figure}

To facilitate autonomy, each lane is designated with a lane identifier $a, b, c, \ldots$, following a clockwise direction starting from true north. These identifiers also signify lane priorities, wherein lane $a$ holds higher priority than lane $b$, and so forth. The significance of lane priorities will be elaborated upon subsequently. A vehicle's intended maneuver is represented by numerical values: 1 for left, 2 for straight, and 3 for right, as depicted in Figure \ref{fig: solution}. Notationally, $V^i_j$ signifies a vehicle in lane $i$ ($i \in {a, b, c, \ldots}$) with an intended maneuver $j$ ($j \in {1, 2, 3}$).

The unanimous decision is required to be optimal in the sense that maximum non-conflicting vehicles can cross at a time. The non-conflicting maneuvers that can be executed at an intersection are immutable, and we capture these non-conflicting maneuvers in the form of an \textit{Harmony Matrix}.
For an intersection $I_n$ with $n$ incoming lanes, the harmony matrix is defined as a $n(n-1)\times n(n-1)$ binary matrix. Each row and column corresponds to maneuver $V_j^i$; $j=1,2,\dots;$ $i = a,b,c,\dots$.  A non-zero entry within the Harmony Matrix indicates that no collision will occur if the respective vehicles execute their intended maneuvers through the intersection simultaneously. For instance, table \ref{table: harmony_mat} demonstrates a harmony matrix for a 4-way intersection similar to one in figure \ref{fig: solution}. For the scenario depicted in figure \ref{fig: solution}, the $V^a_1$ column in the harmony matrix has non-zero values in $V^c_2$ and $V^d_2$ rows. That means $V^a_1$ has harmony with $V^c_2$ and $V^d_2$. Although this doesn't mean $V^c_2$ and $V^d_2$ also have harmony. In a more generalized scenario, the $m^{th}$ column of the harmony matrix indicates the harmony between all possible maneuvers in other lanes independently. The harmony matrix is used as a look-up table for the decision-making algorithm and is stored offline on the ego vehicle.

\begin{table*}
\caption{Harmony matrix for a four-way intersection. In the matrix, 0 stands for conflict and 1 for harmony/coexistence}
\begin{center}
\label{table: harmony_mat}
\setlength{\tabcolsep}{10pt}
%\begin{tabular}{|p{25pt}|p{75pt}|p{115pt}|}
 \begin{tabular}{|c|c|c|c|c|c|c|c|c|c|c|c|c|}
    \hline
    &$V_1^a$ & $V^a_2$ & $V^a_3$ & $V^b_1$ & $V^b_2$ & $V^b_3$ & $V^c_1$ & $V^c_2$ & $V^c_3$ & $V^d_1$ & $V^d_2$ & $V^d_3$ \\
    \hline
    $V^a_1$& 0 & 0 & 0 & 1 & 1 & 1 & 1 & 1 & 0 & 1 & 0 & 1\\
    \hline
    $V^a_2$& 0 & 0 & 0 & 0 & 0 & 0 & 1 & 1 & 0 & 1 & 0 & 0\\
    \hline
    $V^a_3$& 0 & 0 & 0 & 1 & 0 & 0 & 0 & 0 & 1 & 1 & 0 & 0\\
    \hline
    $V^b_1$& 1 & 0 & 1 & 0 & 0 & 0 & 1 & 1 & 1 & 1 & 1 & 0\\
    \hline
    $V^b_2$& 1 & 0 & 0 & 0 & 0 & 0 & 0 & 0 & 0 & 1 & 1 & 0\\
    \hline
    $V^b_3$& 1 & 0 & 0 & 0 & 0 & 0 & 1 & 0 & 0 & 0 & 0 & 1\\
    \hline
    $V^c_1$& 1 & 1 & 0 & 1 & 0 & 1 & 0 & 0 & 0 & 1 & 1 & 1\\
    \hline
    $V^c_2$& 1 & 1 & 0 & 1 & 0 & 0 & 0 & 0 & 0 & 0 & 0 & 0\\
    \hline
    $V^c_3$& 0 & 0 & 1 & 1 & 0 & 0 & 0 & 0 & 0 & 1 & 0 & 0\\
    \hline
    $V^d_1$& 1 & 1 & 1 & 1 & 1 & 0 & 1 & 0 & 1 & 0 & 0 & 0\\
    \hline
    $V^d_2$& 0 & 0 & 0 & 1 & 1 & 0 & 1 & 0 & 0 & 0 & 0 & 0\\
    \hline
    $V^d_3$& 1 & 0 & 0 & 0 & 0 & 1 & 1 & 0 & 0 & 0 & 0 & 0\\
    \hline
\end{tabular}
\end{center}
\end{table*}

The Harmony Matrix facilitates decision-making in $n$-way intersections with $n$ incoming lanes. Each lane contributes up to $n$ vehicles with distinct intents $V^i_j$. To achieve unanimous decision-making and maximize intersection throughput, we construct a graph with at most $n$ nodes representing $V^i_j$ (refer to Figure \ref{fig: solution}). These nodes are interconnected based on Harmony Matrix entries, where edges denote non-zero associations between maneuvers. Identifying the optimal combination entails locating the largest fully connected sub-graph, a classic problem in graph theory known as the Maximal Clique Problem \cite{Frank1986MarkovGraphs}. A clique in an undirected graph denotes a complete subgraph, while a maximal clique is one where no further vertices can be added \cite{Valiente2002CliqueCover}. Solving this problem involves the branch and bound method \cite{Bron1973AlgorithmGraph, Tomita2006TheExperiments}, yielding the largest clique representing maximum vehicles able to traverse the intersection without conflicts, thus optimizing throughput.

The proposed solution is outlined in Algorithm \ref{Pesudo_code}. Upon a vehicle's approach to the intersection, Algorithm \ref{Pesudo_code} is invoked. Initially, the vehicle enters the Red Zone and adjusts its speed to 20 km/h (line 3). Upon entering the Yellow Zone, the vehicle employs onboard sensors to ascertain surrounding vehicles' intended maneuvers through any of the algorithms given in the literature (line 5). Subsequently, a graph is constructed by querying the harmony matrix (line 6). The harmony matrix is a static matrix generated offline and used as a look-up table online for the construction of a graph. The best possible combination of vehicles is determined by solving the maximal clique problem on the created graph (line 7). Various algorithms exist to solve this problem, and in our implementation, we utilize algorithms provided in the \textit{networkx} package. If a single largest clique is identified, the vehicles within it are granted right of way. However, in the presence of multiple cliques of equal size, priority is given to the clique containing vehicles from higher priority lanes (line 8), as determined by their lane IDs $a, b, c, \ldots$. This iterative process continues until the vehicle safely traverses the intersection.

\begin{algorithm}[htbp]
    
    \caption{Intersection management}
    \textbf{Input:} $\mathcal{M}$ $\leftarrow$ \textit{Harmony Matrix}%, \\$IVehs$ $\leftarrow$ \textit{Vehicles and their Maneuvers}
    \begin{algorithmic}[1]
    \While {$vehicle~at~intersection$}
    \If {$vehicle ~ in ~ \textit{Red~Zone}$}
        \State $Reduce~Speed~20km/h$
    \ElsIf {$vehicle ~ in ~ \textit{Yellow~Zone}$}
        \State $IVehs \leftarrow \text{Vehicles and their intent}$
        \State $Graph \leftarrow Create\_graph(IVehs, \mathcal{M})$
        \State $RoW \leftarrow Solve\_max\_clique\_prob(Graph)$
        \State $RoW \leftarrow Select\_priority\_clique(RoW)$
        \If{$Vehicle~in~RoW~and~No~vehicles~crossing$}
        \State $cross~the~intersection$
        \EndIf
    \ElsIf {$vehicle ~ in ~ \textit{Green~Zone}$}
        \State $cross~the~intersection$
    \EndIf
    \EndWhile
 \end{algorithmic}
 \label{Pesudo_code}
 \end{algorithm}

\subsection{Deadlock Analysis}

The proposed algorithm is meticulously crafted to guarantee deadlock-free operation as vehicles navigate through the intersection. Several critical measures are implemented to avert deadlock scenarios.

Primarily, the algorithm is triggered exclusively when a vehicle enters the red zone within the intersection, and it concludes promptly once the vehicle secures the right of way and completes its crossing.

%%%%%%%%%%%%%%%%%%%%%%%%%%%%%%%%%%%%%%%%%%%%%Deadlock proof%%%%%%%%%%%%%%%%
To prove the claim, consider a 4-way intersection, although it is extendable to any other intersection as well. For a 4-way intersection, each lane has four possibilities: absent, left, right, and straight. Hence, there exist $4^4 -1 = 255$ possible cases with at least one vehicle at the intersection. The algorithm constructs a graph with at most 4 nodes using the harmony matrix with the respective vehicles and their maneuver as nodes. Then, the clique with maximal length is considered the optimal solution.
A deadlock occurs when two or more vehicles reach an impasse. This occurs if the vehicles don't have the same solution and they both are of the thought that the other vehicle has the right of way, resulting in an eternal wait. We discuss the two cases below, which can result in deadlock.

\begin{figure}
    \centering
    \includegraphics[width=2.5 in]{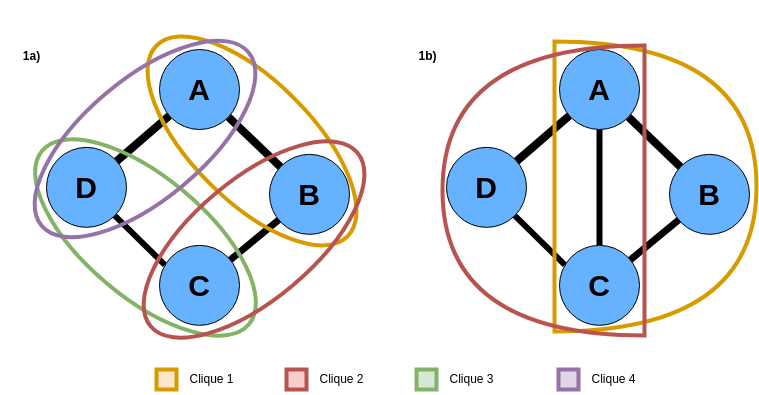}
    \caption{Maximum Cliques possible in a graph with 4 nodes}
    \label{fig:worst-case}
\end{figure}

\textbf{Case 1: Multiple Possibilities of optimal solution}
Let us assume the worst-case scenario of a maximum number of cliques of the same length, i.e., four. The worst case has four maximum number of cliques; the graph in figure \ref{fig:worst-case} explains the possible scenario. The figure \ref{fig:worst-case}a) shows a maximum of four cliques is possible with four nodes, and any further addition of edge results in a lesser maximal clique, as shown in figure \ref{fig:worst-case}b).
Note: The nodes of the graph have only lane ID as a label for the sake of explanation. An example of such a case would be vehicles in A and C intending to go left(1) and B and D intending to go straight(3).

The four cliques given in figure \ref{fig:worst-case}a) are \{[A,B],[B,C],[C,D],[D,A]\}. For the deadlock to occur, let us assume each vehicle chooses a different combination, resulting in deadlock. Now, as per the algorithm, in the case of multiple cliques of the same length, the clique with the highest priority lanes will be selected. Among the 4 cliques \{ [A, B], [D, A] \} contain the highest priority lane 'A,' and after 'A', the highest priority lane is 'B.' Hence, the combination selected by the algorithm is \{[A, B]\}. If the algorithm outputs the combination of [A, B] for all, then our initial assumption that each vehicle chooses a different combination is contradicted. And since all vehicles choose a single combination, i.e., each vehicle knows which vehicles have the right of way, the deadlock never occurs. For instance, Figure \ref{fig: solution} (A) shows a scenario where multiple vehicles have conflicting movements. As per the algorithm, we construct a graph using harmony matrix and vehicle movements as shown in Figure \ref{fig: solution} (B). There are multiple cliques of the same length possible; $(V^c_2, V^a_1)$, $(V^a_1, V^b_2)$, $(V^b_2, V^d_2)$. In this case, the clique with higher lane priority is selected, i.e., $(V^a_1, V^b_2)$.

This is a complete deterministic scenario where only 1 solution exists for every possible case, thus making the probability of a deadlock occurring zero. We have also performed extensive simulations and analyzed all 255 cases to ensure the algorithm is deadlock-free. This result also extends to 3-way, 5-way, and n-way intersections. 
%%%%%%%%%%%%%%%%%%%%%%%%%%%%%%%%%%%%%%%%%%%%%%%%%%%%%%

\textbf{Case 2: Addition of new vehicle after obtaining right of way:}
In rare cases where a new vehicle enters the yellow zone of a previously empty lane after a decision has been made and vehicles with the right of way have started moving, a mismatch in arrival times may occur, potentially jeopardizing unanimous decision-making. To address this, the green zone acts as a safety buffer, preventing vehicles from immediately entering the intersection after a decision is made.
As a result, the algorithm continues running, and when a new vehicle enters the yellow zone post-decision, it is considered in the subsequent iteration, ensuring a new unanimous decision is reached. This mechanism effectively handles edge cases where arrival timing mismatches may disrupt the decision-making process.

If such scenarios occur frequently to the extent that a vehicle is on the verge of entering the intersection, and if more than half of the vehicle's length is within the green zone, it is granted the right of way. Subsequently, the algorithm waits until the intersection is cleared before proceeding further, ensuring safe and efficient navigation through the intersection. The frequency of interventions also decides the length of the green zone, but given that the traffic density is low, the interventions are suspected to be less frequent.

\textbf{Remark 1}: As all the vehicles follow pre-determined non-conflicting paths to cross intersections, safety is assured. In case of discrepancies, a safety monitor can be used to enhance safety. \cite{Tian2022SafetyModels} demonstrate the successful application of such a monitor for merging on roundabouts, which can be extended in this case.

\section{Simulation Setting}
\label{sec:simultaion}
The efficacy of the developed algorithm is assessed by conducting comprehensive simulations replicating real-world scenarios. To simulate realistic traffic conditions, we employ the Simulation of Urban MObility (SUMO) \cite{Lopez2018MicroscopicSUMO}. SUMO is an open-source traffic simulator renowned for its ability to handle large-scale traffic simulations. Within SUMO, the built-in functions are leveraged to facilitate motion planning, enabling seamless evaluation of intersection management algorithms. To facilitate real-time control of vehicles, the Traffic Control Interface (TraCI), a Python API specifically designed for SUMO, is utilized.

The developed algorithm is subjected to a simulation duration of 1 hour under conditions of low to medium traffic density. To determine the appropriate range for low to medium traffic density, we refer to the Manual on Uniform Traffic Control Devices (MUTCD), specifically Chapter 4 \cite{2009ManualHighways}, which states that a traffic signal is warranted if the traffic volume on a single-lane road reaches 500 PCUs per hour or above for both directions. Thus, we conduct simulations across various values: 150, 200, 250, 300, and 350 PCUs/hour/lane.

\begin{figure}   
\centering
\includegraphics[width = 3  in]{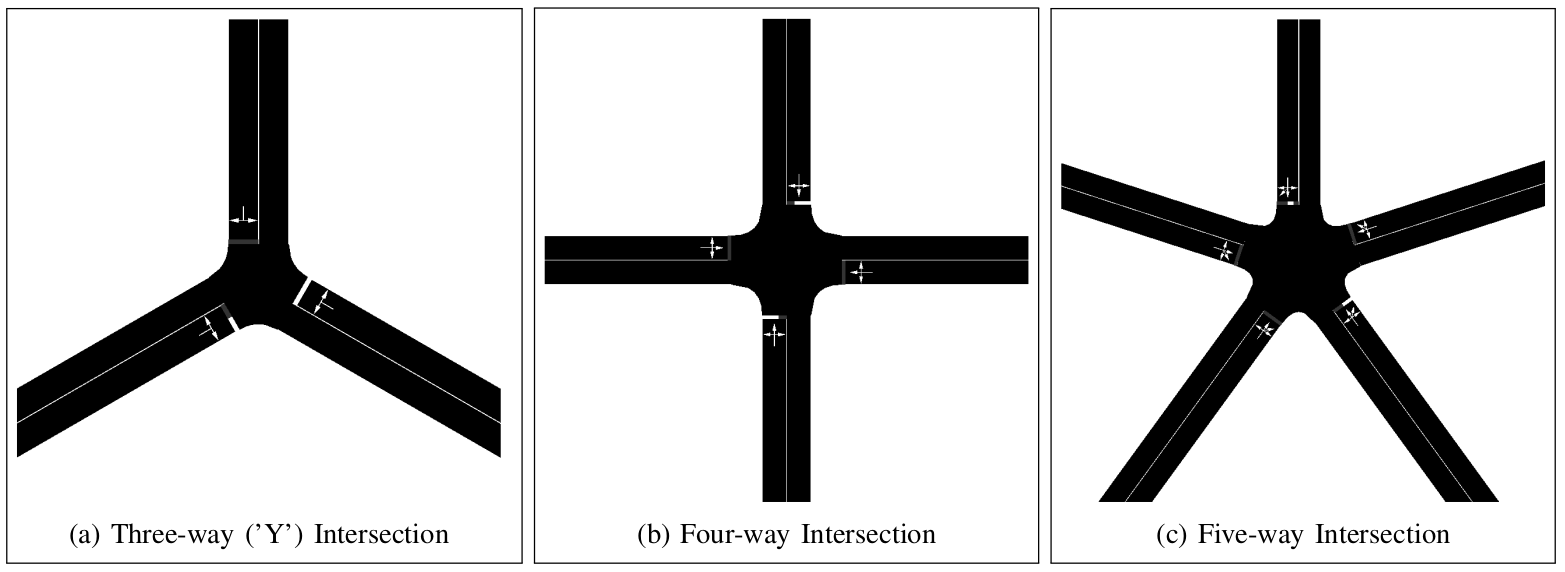}
\caption{Various intersections considered for analysis}
\label{fig:Maps}
\end{figure}

\textbf{\emph{Intersection Types}}:
The algorithm's robustness is evaluated on various types of intersections, including a three-way intersection with a 'Y' shape, a four-way junction, and a five-way junction. These intersections are characterized by roads extending up to 500m, each with a single incoming and outgoing lane. Figure \ref{fig:Maps} provides a visual representation of the three junctions that are considered for testing the algorithm's performance.

\textbf{\emph{Real-world traffic modeling/distribution}}:
Assuming a uniform distribution of incoming vehicles is unrealistic in real-world scenarios. To model the arrival of vehicles more accurately, we employ a \textit{Poisson Distribution}. The Poisson Distribution, represented by Equation \ref{eqn: poisson}, describes the probability of an event occurring a certain number of times within a fixed time interval. 
\begin{equation}
    P(k)=\frac{\lambda^ke^-\lambda}{k!}
\label{eqn: poisson}
\end{equation}

In the equation, $P(k)$ denotes the probability of the event occurring $k$ times, while $\lambda$ represents the average number of events in the fixed time interval $t$. The Poisson distribution is particularly useful for modeling independent and random events, given the knowledge of the average occurrences within a specific time interval. In our context, we utilize the Poisson distribution to estimate the probability of incoming vehicles. The $\lambda$ is the function of volume density per lane, ranging from 150 to 350 PCUs/hr/lane. These values serve as the average rate of events in the Poisson distribution, allowing us to calculate the probability of different numbers of vehicles arriving in a given time period.

\section{Results and Discussion}
\subsection{Evaluation Metrics}
The intersection management algorithm's performance is assessed using two key metrics: travel time and average waiting time. These metrics offer a comprehensive evaluation of the algorithm's impact on traffic flow, congestion, and overall intersection efficiency.

\subsubsection{Travel Time}
Travel time is defined as the duration it takes for a vehicle to traverse a road segment extending 500 meters from the center of the intersection on each side. It encompasses the time from when a vehicle arrives at the intersection until it completely passes through the specified road segment.

\subsubsection{Average Waiting Time}
Average waiting time measures the duration that vehicles spend in a queue at the intersection before being able to proceed. It represents the average time a vehicle waits before entering the intersection.

\subsection{Comparison Models Reasoning}
To assess the algorithm's effectiveness, we perform a comparative study using non-communicative and communicative methods defined below. The comparative study consists of two types of traffic distribution: 1. Balanced traffic and 2. Unbalance traffic. In balanced traffic, the traffic density is equal in all lanes, and in unbalanced traffic, the traffic densities are distributed non-uniformly, and the lane priorities are set accordingly.
\subsubsection{Fixed-Time Traffic Signal (FTS)}
A fixed-time traffic signal is the most common method to control junction traffic. In a fixed-time traffic signal, the green time is fixed and does not change with respect to time. Webster's formula \cite{V.1958TrafficSetting} is often used to determine the optimal cycle length and effective green time. 
\begin{equation}
    \textrm{Optimal cycle length} (Co)=\frac{1.5*L+5}{1-y}
\label{eqn: Webster_Co}
\end{equation}
\begin{equation}
    \textrm{Effective Green Time}(Ga)=\frac{ya/y}{Co-L}
\label{eqn: Webster_Ga}
\end{equation}

\noindent where $L$ represents the total lost time, including all red time, we set $L=2n$, where $n$ is the number of incoming lanes. $y$ is the sum of critical ratios for each lane, which is the ratio of observed volume to saturation flow.
Saturation flow refers to the maximum number of PCUs that can pass per hour. To establish the saturation flow, we refer to an empirical study by \cite{VasanthaKumar2018StudyIndia}. In their study, they observed a busy three-legged junction in Vellore, India, and noted a peak traffic of 7573 PCUs per hour at the junction. Since the three-legged junction has 5 incoming lanes, we consider a saturation flow of approximately $7573/5 \approx 1500$ in our case.
Table \ref{table: sim_tab} displays phasing timings for the respective observed volumes. During the simulation, we utilize a 4-phase traffic signaling approach since there is a single incoming lane, and all movements are required during the green time.

\subsubsection{Adaptive Traffic Signal (ATS)}
Adaptive traffic signals dynamically adjust to traffic demands, enhancing traffic flow. Real-time data is gathered through traffic sensors or cameras, and algorithms then adapt signal timings according to this data. We employ a delay-based algorithm developed by \cite{Oertel2011Delay-TimeIntersection} for comparison. This algorithm modifies green times based on queue sizes, adhering to maximum and minimum green time bounds. In practice, the algorithm is given fixed-time signals as per Table \ref{table: sim_tab}, and then the algorithm modifies them.

\begin{table}
\caption{Signal phase timing for fixed-time signal}
\begin{center}
\label{table: sim_tab}
\setlength{\tabcolsep}{10pt}
%\begin{tabular}{|p{25pt}|p{75pt}|p{115pt}|}
 \begin{tabular}{|c|c|c|}
    \hline
    Traffic Density & Green time(s) & Amber time(s) \\
    \hline
    150 & 5 & 2\\
    \hline
    200 & 7 & 2\\
    \hline
    250 & 10.75 & 2\\
    \hline
    300 & 19.75 & 2\\
    \hline
    350 & 61.75 & 2\\
    \hline
\end{tabular}
\end{center}
\end{table}

%%%%%%%%%%%%%%%%%%%%%%%%%%%%%%%%%%%%%%%%%%%%%
\subsubsection{Intersection Management using V2I protocols (V2I-C)}
We also compare our strategy and an algorithm that employs communication for intersection management. \cite{Li2020IntersectionCommunication} utilize vehicle-to-infrastructure communication to gather data. Using this collected data, conflicts are identified through a predefined conflict matrix, and the infrastructure provides vehicle arrival times. Vehicles then adapt their speeds based on these arrival times, ensuring a seamless passage through the intersection.

\begin{figure}
    \centering
    \includegraphics[width = 2.5  in]{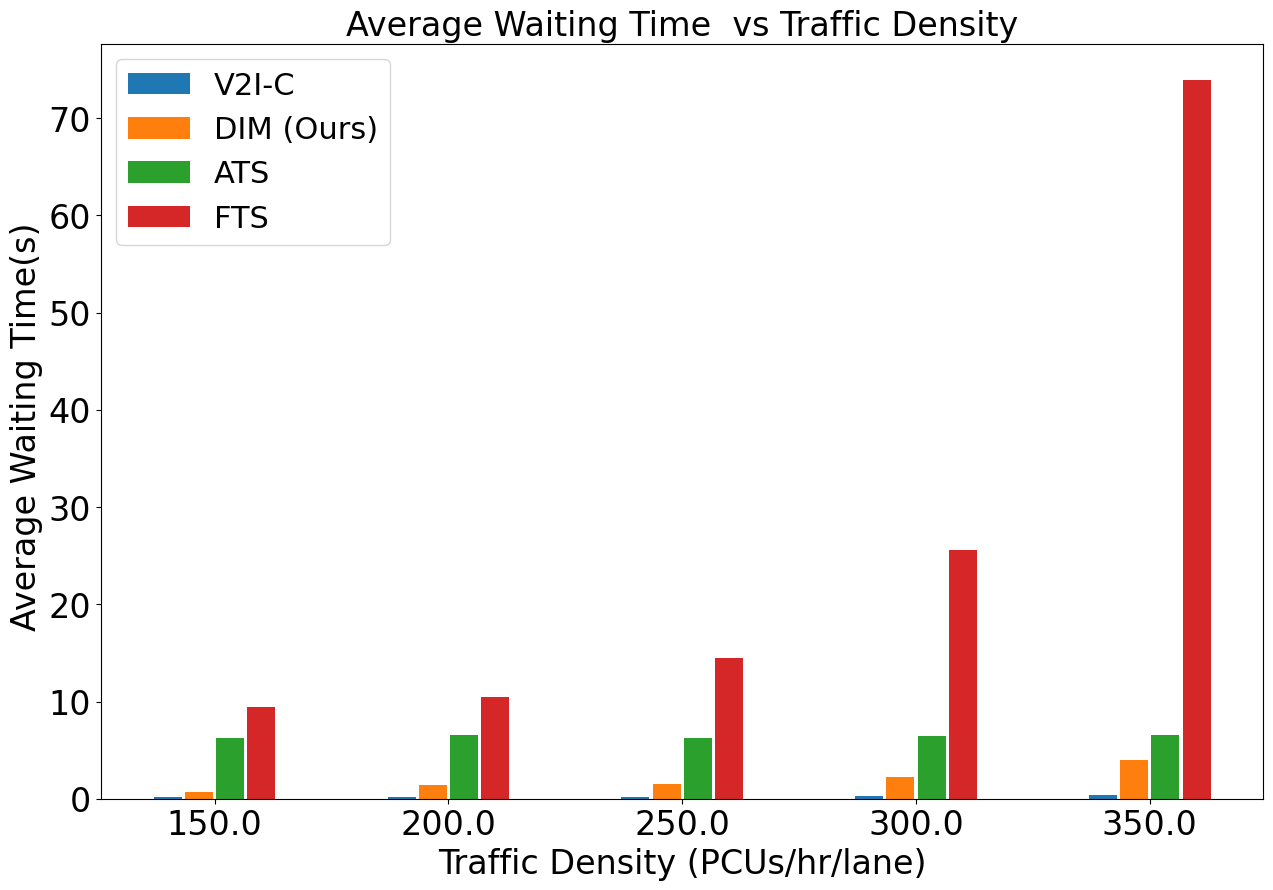}
    \caption{Comparative study on average Waiting Time(s) vs Traffic Density (PCUs/hour/lane) for balanced traffic}
    \label{fig: glob_WT}
\end{figure}

\begin{figure}
    \centering
    \includegraphics[width = 2.5  in]{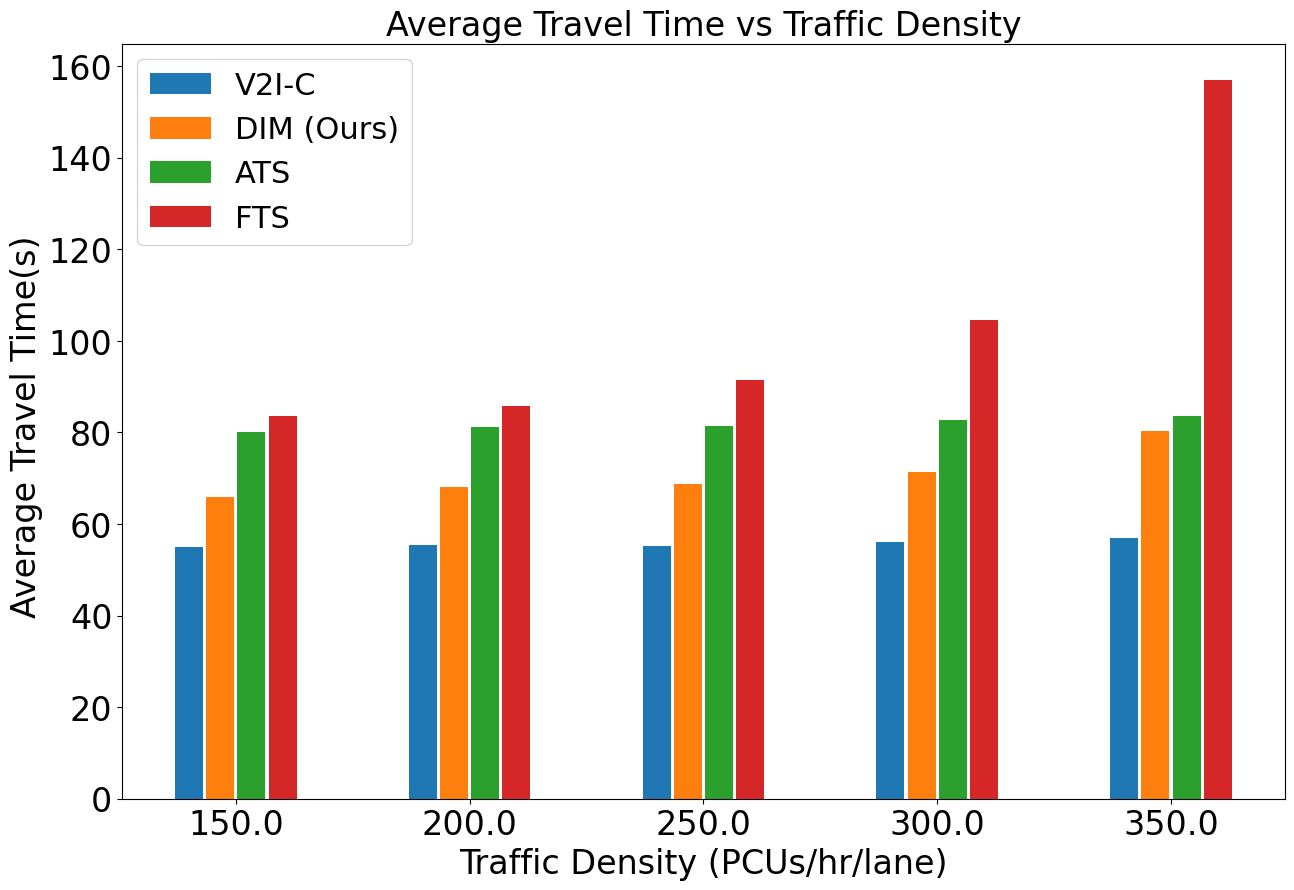}
    \caption{Comparative study on average Travel Time(s) vs Traffic Density (PCUs/hour/lane) for balanced traffic}
    \label{fig: glob_TT}
\end{figure}

Figure \ref{fig: glob_WT} and \ref{fig: glob_TT} show the comparison study between the 4 algorithms for Balanced traffic. The waiting time delay of our approach is lower than fixed-time traffic signals (FTS) and adaptive traffic signals (ATS) and comparable to V2I-C's at lower traffic densities. A similar trend is observed in travel time. At a traffic density of 350 PCUs/hr/lane, V2I-C's waiting time and travel time are 11.8 and 1.4 times lower than ours, respectively, but considering the fact that they have used V2I communication for management, it is expected. The huge difference in the waiting time is because, in V2I-C, the vehicles slow down before approaching the intersection so that the vehicles do not stop at the intersection.  On the contrary, in our Decentralized Intersection Management(DIM) algorithm, the vehicles make a stop at the intersection, leading to a reasonable waiting time. In this context, comparing travel times offers more accurate comparisons, and our algorithm yields comparable results at low traffic densities without using any infrastructure.

\begin{figure}
    \centering
    \includegraphics[width =2.5 in]{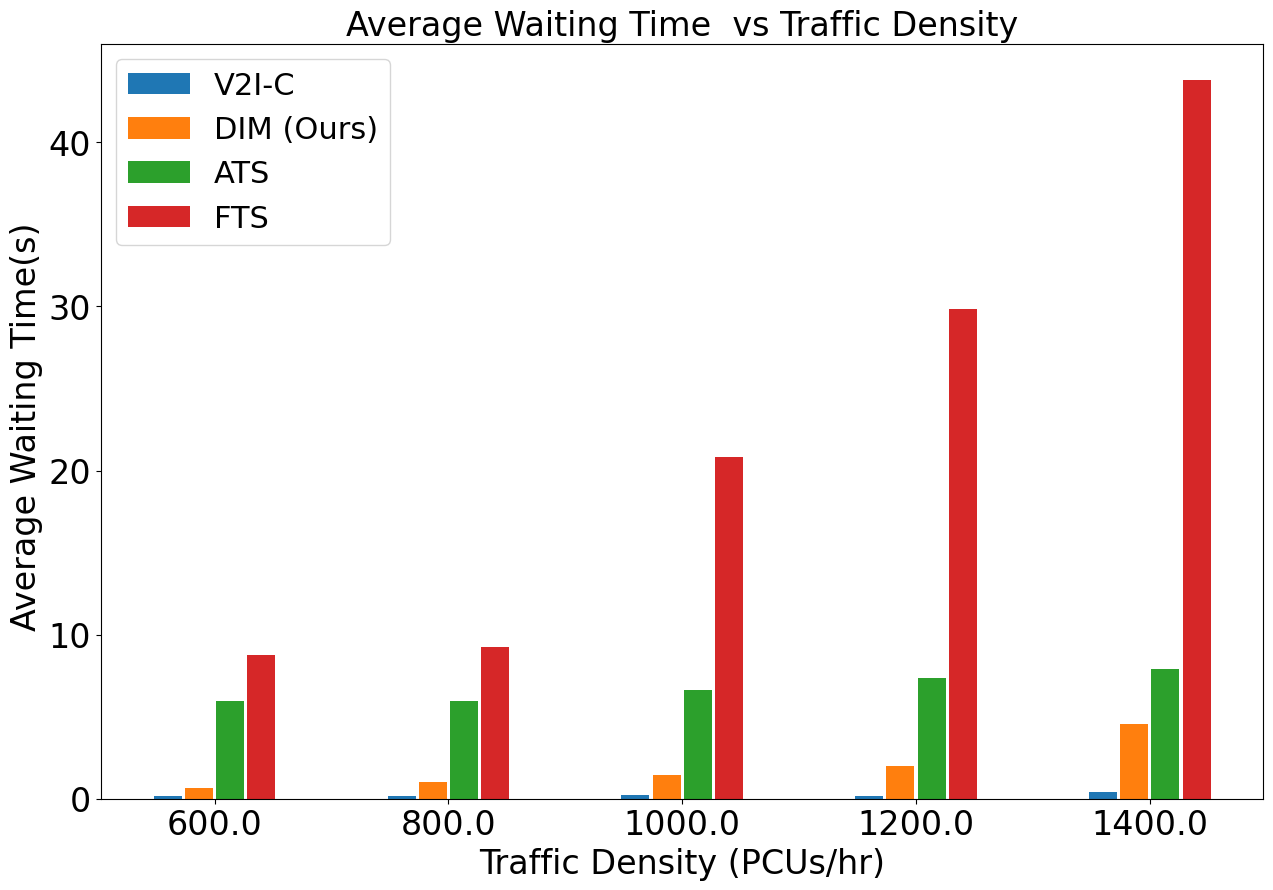}
    \caption{Comparative study on average Waiting Time(s) vs Traffic Density (PCUs/hour/lane) for unbalanced traffic ratio of 4:3:2:1}
    \label{fig: un_WT}
\end{figure}

\begin{figure}
    \centering
    \includegraphics[width = 2.5  in]{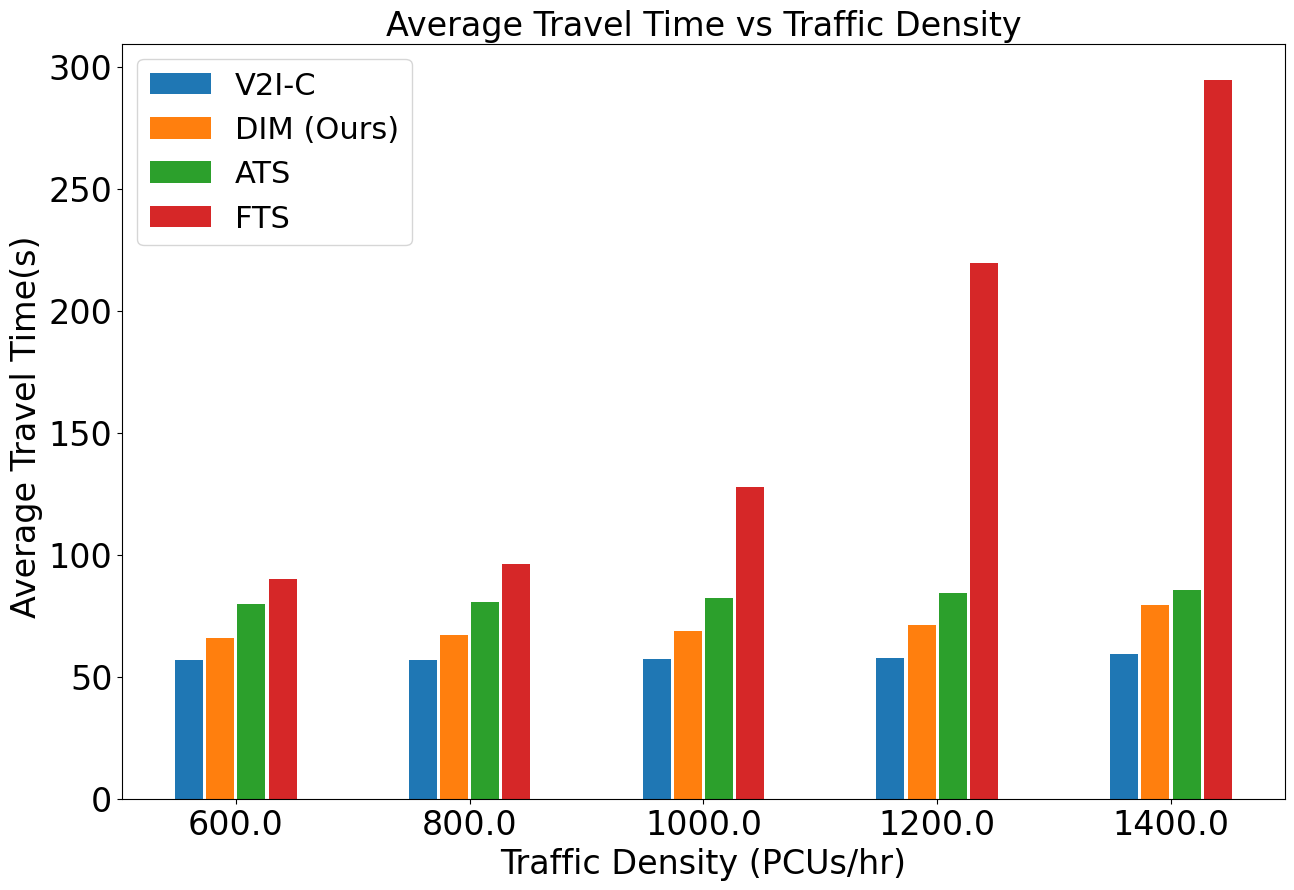}
    \caption{Comparative study on average Travel Time(s) vs Traffic Density (PCUs/hour/lane) for unbalanced traffic ratio of 4:3:2:1}
    \label{fig: un_TT}
\end{figure}

In unbalanced traffic density, the total traffic density is divided into the ratio of 4:3:2:1 and 4:1:4:1, and the ratios denote the volume of traffic incoming from North:East:South:West. The lane priorities are set according to the incoming traffic, the highest being the one with the maximum incoming traffic. Figure \ref{fig: un_WT} and \ref{fig: un_TT} show the average waiting time and average travel time for unbalanced traffic. The performance of fixed-time and adaptive traffic signals declines, whereas our algorithm's performance is improved compared to balanced traffic. This is due to the fact that we use absolute lane priorities, which results in queuing in the least priority lanes for balanced traffic, whereas the queue length in unbalanced traffic is lower, and a large amount of traffic is resolved faster by assigning higher priority to the lane. The highest average travel time is within the bound of 50\% of the V2I-C algorithm with communication infrastructure.

\begin{figure}
    \centering
    \includegraphics[width = 2.5  in]{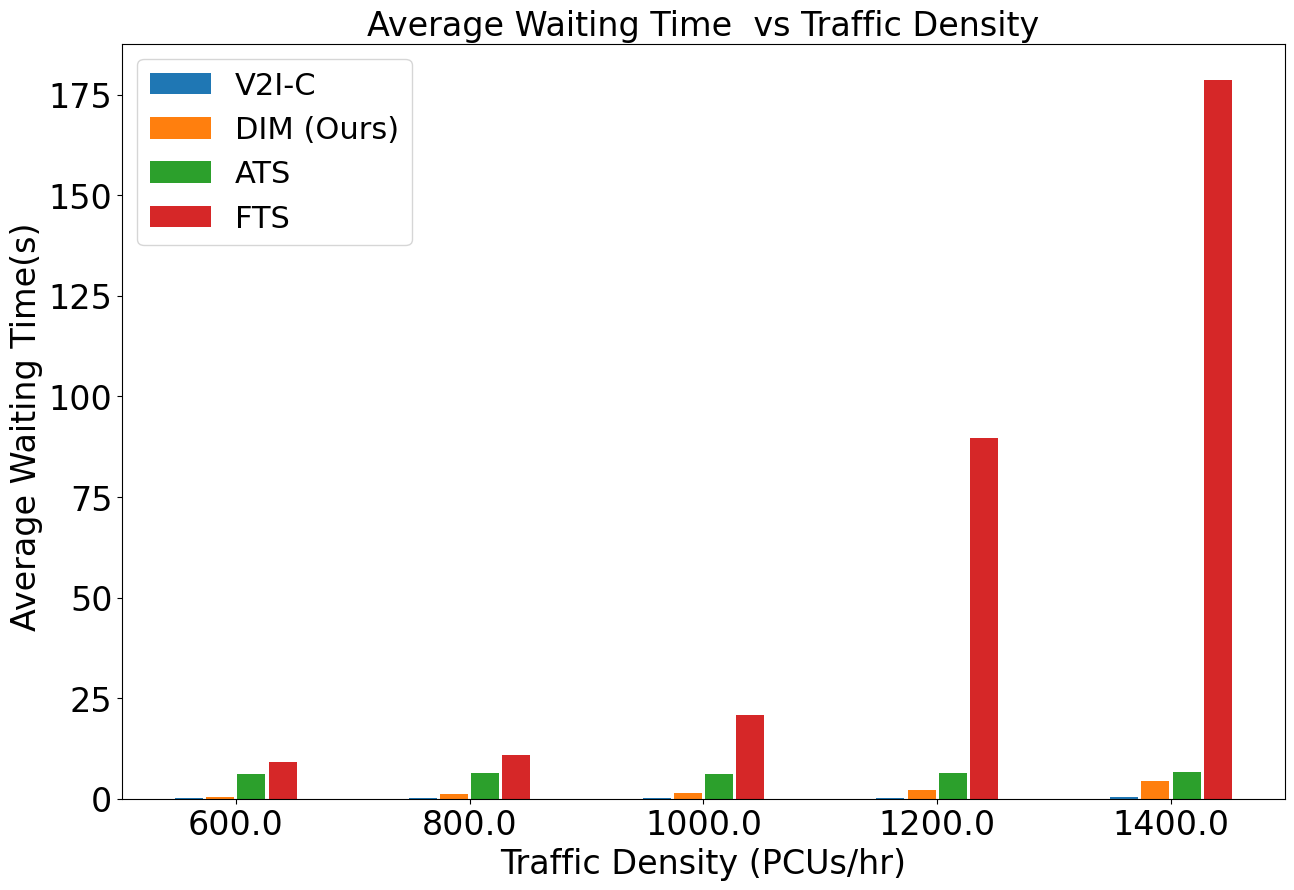}
    \caption{Comparative study on average Waiting Time(s) vs Traffic Density (PCUs/hour/lane) for unbalanced traffic ratio of 4:1:4:1}
    \label{fig: un_WT_41}
\end{figure}

\begin{figure}
    \centering
    \includegraphics[width = 2.5  in]{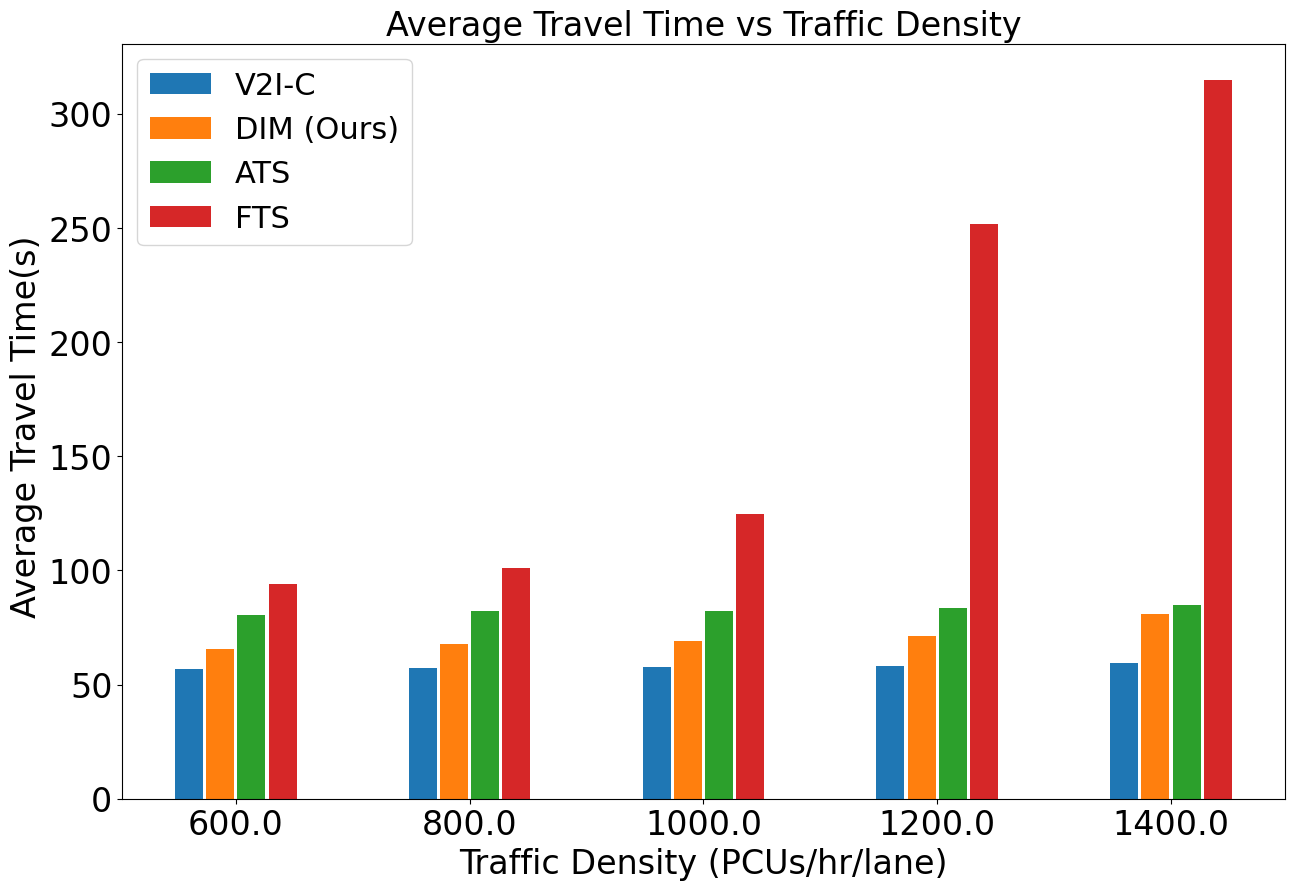}
    \caption{Comparative study on average Travel Time(s) vs Traffic Density (PCUs/hour/lane) for unbalanced traffic ratio of 4:1:4:1}
    \label{fig: un_TT_41}
\end{figure}

Figure \ref{fig: un_WT_41} and \ref{fig: un_TT_41} depict the results for unbalanced traffic with a ratio of 4:1:4:1. The results show a similar trend to unbalanced traffic with a ratio of 4:3:2:1.
%%%%%%%%%%%%%%%%%%%%%%%%%%%%%%%%%%%%%%%%%%%%%%%%%%
\subsection{Intersections types}
\begin{figure}
    \centering
    \includegraphics[width = 2.5  in]{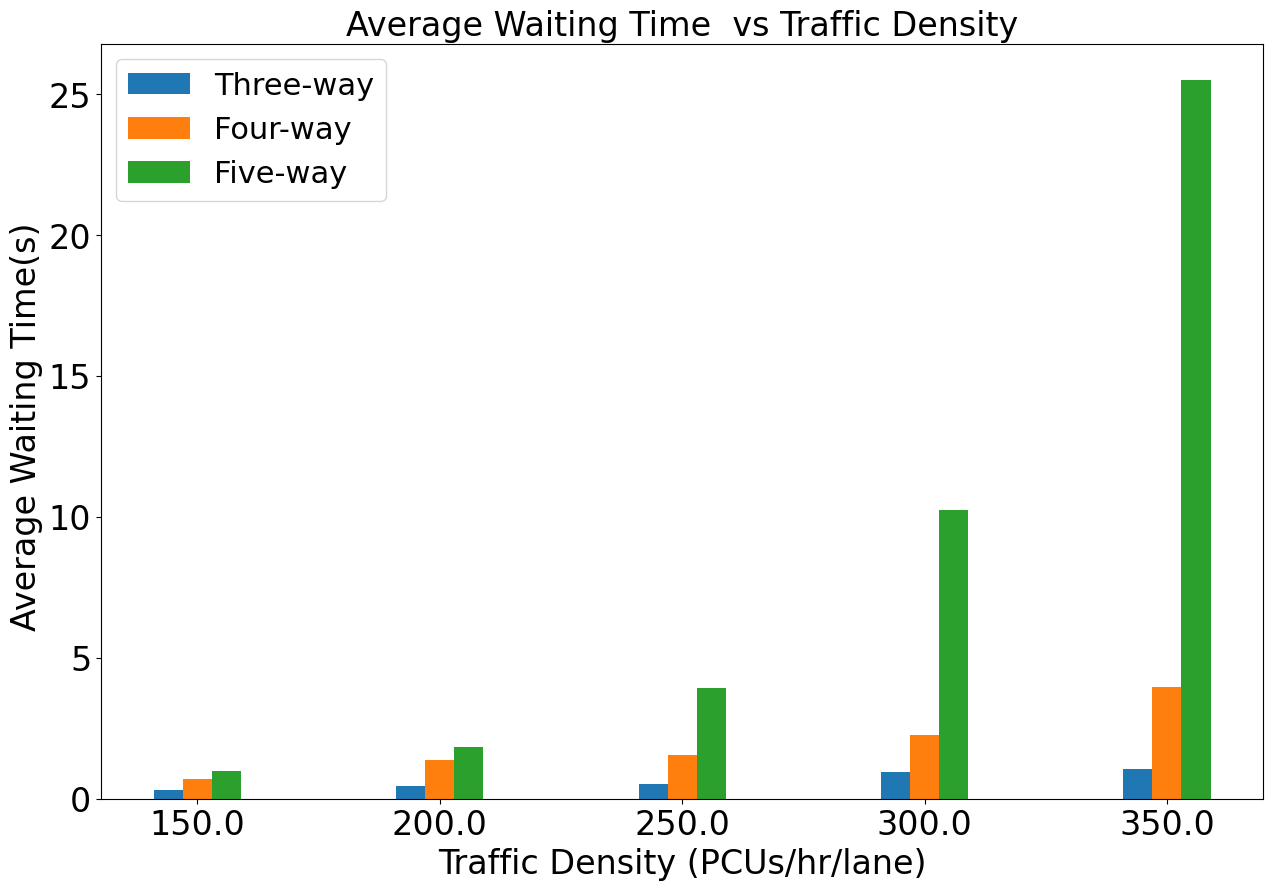}
    \caption{Average Waiting Time(s) vs Traffic Density (PCUs/hour/lane) for different intersections}
    \label{fig: WT}
\end{figure}

\begin{figure}
    \centering
    \includegraphics[width = 2.5  in]{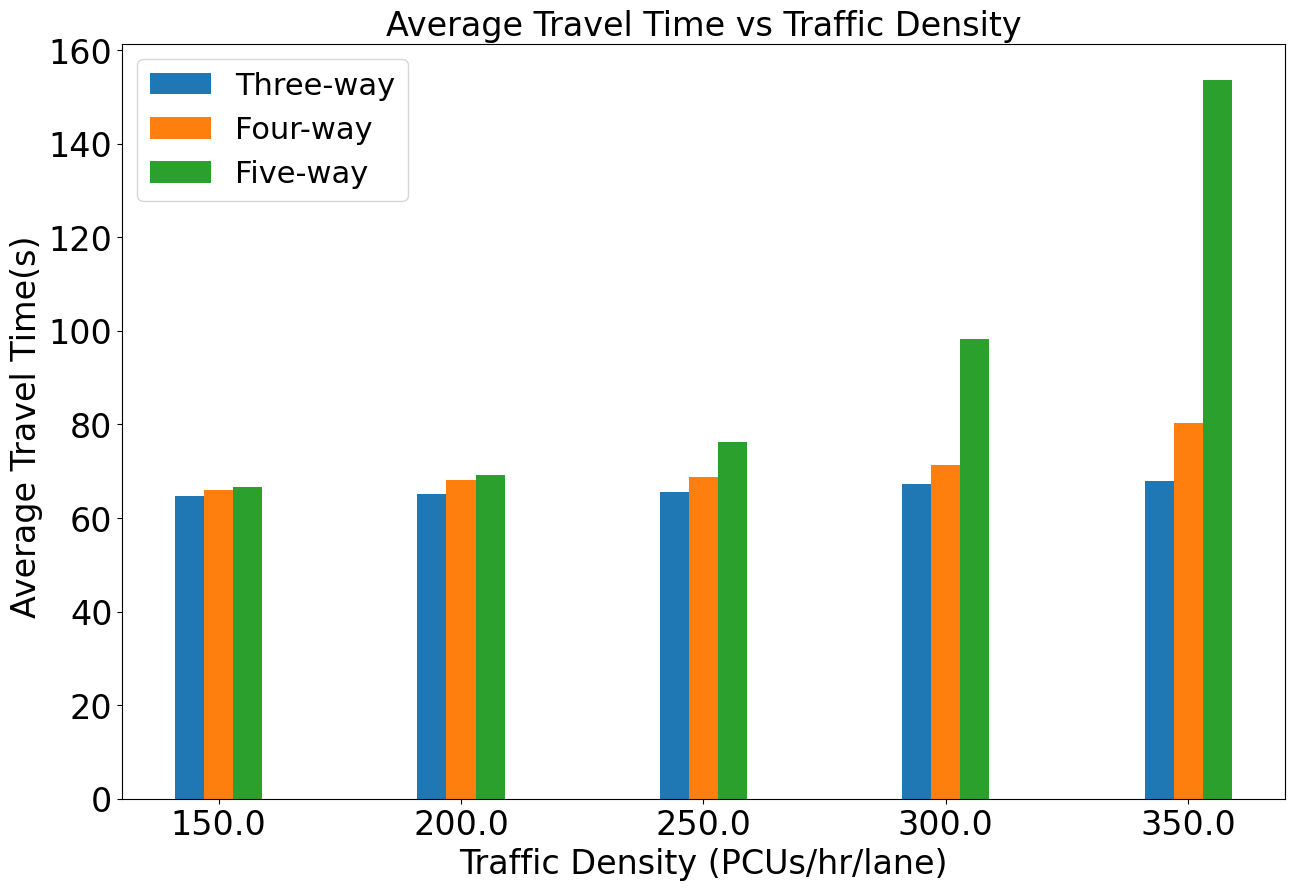}
    \caption{Average Travel Time(s) vs Traffic Density (PCUs/hour/lane) for different intersections}
    \label{fig: TT}
\end{figure}

The versatility of the algorithm allows for seamless extension to intersections with $n$ number of incoming lanes by adjusting the conflict matrix. Here, we showcase the algorithm's performance across 3-way, 4-way, and 5-way intersections. Figures \ref{fig: WT} and \ref{fig: TT} illustrate the average waiting time and average travel time for varying traffic densities, respectively.

The average waiting time increases with the number of incoming lanes as the traffic density is directly proportional to incoming lanes. The average time delay does not vary as much as the average waiting time but shows a similar trend. This is because the queue length in the least priority lane increases, leading to a higher waiting time for vehicles in the lane. Table \ref{table: lane_delay} shows the average waiting time for each lane at a traffic density of 350 PCUs/hr/lane. It can be clearly observed that the values for the least priority lane (alphabetical priority is used as defined in problem formulation) values increase significantly.

\begin{table}
\caption{Lane-wise waiting time (s) for 3-way, 4-way, 5-way intersections at a traffic density of 350 PCUs/hr/lane}
\begin{center}
\label{table: lane_delay}
\setlength{\tabcolsep}{10pt}
%\begin{tabular}{|p{25pt}|p{75pt}|p{115pt}|}
 \begin{tabular}{|c|c|c|c|}
    \hline
    Traffic Density & 3-way & 4-way & 5-way \\
    \hline
    lane 'a` & 0.4606 & 0.8660 & 1.0970\\
    \hline
    lane 'b` & 0.9275 & 1.7263 & 2.0858\\
    \hline
    lane 'c` & 1.6415 & 2.8863 & 4.2837\\
    \hline
    lane 'd` & N/A & 10.2094 & 13.3636\\
    \hline
    lane 'e` & N/A & N/A & 146.1092\\
    \hline
\end{tabular}
\end{center}
\end{table}

\section{Conclusion and Future works}
In this study, we introduced a low-cost framework enabling fully autonomous vehicles to navigate unsignalized intersections in low-traffic density scenarios. The algorithm optimizes intersection throughput without relying on dedicated infrastructure. Modifications to the current road system regarding lane color codes and priorities were established, incurring minimal implementation costs for these markers.

Evaluation conducted through traffic simulations in SUMO showcased the algorithm's superior performance compared to existing non-communicative methods (FTS and ATS) and competitive results relative to a communicative approach (\cite{Li2020IntersectionCommunication}). Particularly effective in unbalanced traffic environments, the algorithm's prioritization mechanism renders it well-suited for uncontrolled intersections, minimizing unnecessary infrastructure expenditure in low-traffic density areas.

In our future work, we aim to achieve similar results for mixed autonomy traffic. A rule-based method similar to \cite{Aksjonov2021Rule-BasedEnvironment} to accommodate human drivers is an interesting direction to be explored. Along with this, we would like to use confidence modeling to avert the discrepancies in the intent given by the other algorithms, making the approach robust to noisy intents.

%%%%%%%%%%%%%%%%%%%%%%%%%%%%  THE END  %%%%%%%%%%%%%%%%%%%%%%%%%%%%%%%%%%%%%
%%%%%%%%%%%%%%%%%%%%%%%%%%%% THANK YOU %%%%%%%%%%%%%%%%%%%%%%%%%%%%%%%%%%%%%
\bibliographystyle{unsrt}
\bibliography{main}
\end{document}